\pdfoutput=1

\documentclass[11pt]{article}
\usepackage{authblk}
\usepackage{acl}

\usepackage{times}
\usepackage{latexsym}

\usepackage[T1]{fontenc}

\usepackage[utf8]{inputenc}

\usepackage{microtype}
\usepackage{amsfonts}
\usepackage{microtype}
\usepackage{graphicx}
\usepackage{amsmath}
\usepackage{mathbbol}
\usepackage{amssymb}
\usepackage{xcolor} 

\title{Make The Most of Prior Data: A Solution for Interactive Text Summarization with Preference Feedback}

\author{Duy-Hung Nguyen$^{1,*}$, Nguyen Viet Dung Nghiem$^{1,*}$, Bao-Sinh Nguyen$^1$, Dung Tien Le$^1$, \authorcr Shahab Sabahi$^1$, Minh-Tien Nguyen$^{1,2}$, Hung Le$^3$ \\
        $^1$Cinnamon AI, 10th floor, Geleximco building, 36 Hoang Cau, Dong Da, Hanoi, Vietnam. \\
        \texttt{\{\normalsize hector, henry93, simon, nathan, sshahab, ryan.nguyen\}@cinnamon.is} \\
        $^2$Hung Yen University of Technology and Education, Hung Yen, Vietnam. \\
        \texttt{\normalsize tiennm@utehy.edu.vn} \\
        $^3$Deakin University, Australia. \\
        \texttt{\normalsize thai.le@deakin.edu.au}}

\begin{document}
\maketitle               
\def\thefootnote{*}\footnotetext{These authors contributed equally to this work.}

\begin{abstract}
For summarization, human preferences is critical to tame outputs of the summarizer in favor of human interests, as ground-truth summaries are scarce and ambiguous. Practical settings require dynamic exchanges between humans and AI agents wherein feedback is provided in an online manner, a few at a time.  In this paper, we introduce a new framework to train summarization models with preference feedback interactively. By properly leveraging offline data and a novel reward model, we improve the performance regarding ROUGE scores and sample-efficiency. Our experiments on three various datasets confirm the benefit of the proposed framework in active, few-shot and online settings of preference learning.  
\end{abstract}
%
%
%

%

\section{Introduction}



    
    
    
        

  The advent of AI has changed business practices, though the human involvement is still important. The human roles in interaction with AI-powered machines have been evolving under the concept of \textit{human-in-the-loop (HITL)} \cite{zanzotto2019human}. HITL allows humans to actively participate in supervising AI systems by approving, rejecting, or re-labeling current outputs, and providing expert-guided advices to the system. It will also act as the unique source of external knowledge from humans. By observing the outputs of AI systems, humans can hand-pick some potential outcomes and then feedback to the models for better performance. 


In NLP, document summarization is considered as a subjective task \cite{stiennon2020learning}. They argued that it would be hard to quantify what makes a “good summary” without the human judgment input. Collecting human feedback and evaluating the crafted summaries from documents for building the training datasets is time-consuming \cite{wu2021survey} and labor-expensive. It is true particularly where the domain knowledge is required. Moreover, \citet{uccetina2021survey} argued the importance of the user’s intentions for modeling Natural Language Understanding with high performance.  

Given these attributions, we are more interested in deploying an \textit{interactive} HITL-based text summarization framework, which continuously collects the user-feedback to consequently improve model prediction robustness.
Here, the user’s intention is implicitly acknowledged as a factor influencing the extraction of important sentences from the source documents.  Upon this formulation, the AI model will be trained with human-produced summaries and adapted as more human-feedback is fed in.

Previous studies used human feedback to rank the label of objects \citep{JMLR:v18:16-634}, which employed reinforcement learning to minimize the user's effort to provide feedback for training a ranker. 
However, these approaches rank the entire solution space including relevant and irrelevant pairs, which are a waste of computing power. 
To tackle this shortcoming, \citet{siddhant2018deep} employed Bayesian Optimization, which substitutes the standard uncertainty-based acquisition functions for active learning. However, the model still consumes a lot of computing power for a larger number of iterations, and is vulnerable to the curse of dimensionality of input data. Recently, researchers proposed to learn a reward model simulating human preferences \cite{ziegler2019fine}. The reward model was then used to translate real-time human feedback to the reward score for fine-tuning the model under RL training. The method was designed for online learning of language models and required numerous interactions to achieve good performance. However, how to achieve sample-efficiency is still an open question. 

In this paper, we propose a novel interactive preference learning for the summarization task. To do that, we fine-tune a pretrained extractive summarizer as the backbone with reinforcement learning by using a reward model that enforces the distance-based order of preferences. The reward model is trained to differentiate two summaries regarding the topic, length, and quality (human preferences). To enable sample-efficiency, we propose to utilize \textit{offline} data, which was previously used to pretrain the model. We show that naively using the offline data is harmful for preference learning. Instead, we introduce two mechanisms to selectively sample offline data in favor of human feedback learning. Our sampling strategies focus on low-rewarded samples or documents which are similar to fine-tuning data. We demonstrate that our method can be used in various settings: active, few-shot and online learning. Tested on three summarization datasets, our method consistently achieves significantly better results compared to competitive baselines in each setting. In summary, our contribution is three-fold:
\begin{itemize}
\item We propose a new RL-based preference learning system for the summarization task by using a novel reward model.

\item We propose sampling mechanisms to efficiently leverage offline data for preference learning of extractive summarization.

\item We conduct extensive empirical studies on three summarization datasets, showing that our proposed method outperforms competitive baselines in various settings such as active, online, and few-shot learning scenarios. 
\end{itemize}

\section{Related Work}

\subsection{Summarization with RL}

\paragraph{Direct reward}

The most direct way to have a reward function in reinforcement learning for summarization is to match the candidate summary to the reference (gold) summary \cite{narayan2018ranking,paulus2017deep}. Put in the HITL setting, humans are required to provide the gold summary for training the machine, which is prohibitively expensive and can be ambiguous. Thus, we focus on rewards constructed from preference feedback, in which humans only need to indicate the better summary between two candidate summaries.

\paragraph{Preference reward}

As stated in the introduction, the preference from humans is much more accessible and consistent. In this approach, the frameworks \cite{gao2018april,stiennon2020learning, nguyen2021robust} consist of two main steps: 1) Preference learning that gives a score, which mimics human evaluation, to a summary of a document. 2) Reinforcement learning based on the reward model. These works have not examined the summarization problem in an interactive training scheme, which will be addressed in this paper.

\subsection{Preference learning in NLP}

Preference learning aims at obtaining the ranking (i.e. total ordering) of objects from pairwise
preferences, in which the linear Bradley-Terry (BT) model \cite{bradley1952rank} is one of the most studied
methods.
Later, the APRIL framework \cite{gao2018april} shows that it can reduce the number of required comparisons by using active selection with uncertainty sampling \cite{avinesh2017joint}. 
On the other hand, the OpenAI framework \cite{stiennon2020learning} uses a neural structure to predict the score given a document and its summary. Its advantage is the learning with direct human evaluation (score for each summary) or human preferences (comparison between 2 summaries). However, the input space is the product of two documents, so it needs much more human feedback to achieve good performance. In contrast, our proposed reward model treats the input as only one document, and thus is simpler and has fewer parameters in the learning process. That is more suitable for the interactive learning context where the number of training samples is limited.  

Reinforcement learning is a popular and effective approach to utilize human preferences. The APRIL framework \cite{gao2018april} trains a summarizer by using a pretrained preference model. The work of \citet{ziegler2019fine} employs preference-based RL to fine-tune a deep language model for sentiment classification and document summarization tasks. They only rely on online data during fine-tuning. We instead make use of offline data to accelerate the learning process.

\section{Background}
\subsection{The backbone model} \label{backbone}
We use BERTSUM \cite{Liu-BERTSum-EMNLP-19} as the backbone for extractive summarization. Given a document with a set of sentences, the model uses BERT \cite{devlin2018bert} for learning hidden vectors of sentences by using the modified [CLS] token of each sentence. The vectors are fed into an inter-sentence layer by using the Transformer for learning the inter-relationship among sentences. Important sentences are extracted by using a sigmoid function for sentence importance estimation.




\subsection{Fine-tuning with reinforcement learning}\label{FRL}

\paragraph{Proximal policy optimization}(PPO)
To fine-tune the backbone with interactive feedback, we treat the summarization process as a sequential decision-making process so that we can employ RL \cite{stiennon2020learning, ziegler2019fine}. The RL agent traverses all sentences in the original document, and at each time step, it classifies the current sentence $s_i$ into two labels: important ($y_i=1$) and unimportant ($y_i=0$).
We consider the backbone model trained with supervised learning as the initial policy. The policy is then optimized by using PPO \cite{schulman2017proximal} as our RL method. The objective of training PPO is:
\begin{equation*}
\begin{split}
    J^{PPO}(\theta) = \mathop{\mathbb{E}}[\min(ra(\theta)\hat{A}_{\theta_{old}}(s_i,y_i),\\
    \mathrm{clip}(ra(\theta),1-\epsilon,1+\epsilon)\hat{A}_{\theta_{old}}(s_i,y_i))]
\end{split}
\end{equation*}
where $\theta$ is the current policy's parameters, $\hat{A}_{\theta_{old}}(s_i,y_i)$ is the advantage calculated at the old policy parameters $\theta_{old}$ before each updated policy iteration, by using any advantage estimation algorithm to transform the rewards \cite{schulman2015high}, and $ra(\theta) = \frac{\pi_{\theta}(y_i|s_i)}{\pi_{\theta_{old}}(y_i|s_i)}$ is the ratio between the new policy and the old policy. This is the idea of importance sampling that evaluates the new policy with samples collected from the older policy. If the ratio $ra$ falls outside the range $1-\epsilon$ and $1+\epsilon$, the advantage function will be clipped. PPO uses the objective to avoid big changes between new and old policies.




\paragraph{Reward schemes}
Besides the final reward that evaluates the quality of the whole summary,  we follow \citet{pasunuru2018multi,li2019deep} to use an additional reward to constrain policy updates. Let $\pi_{\theta_0}$ denotes the supervised trained backbone model and $\pi_{\theta}$ is the one that we optimize with RL. The reward at the time step $i$ is:

\begin{equation*}
\begin{split}
    r_i = -\beta_{KL}\log{[\frac{\pi_{\theta} (y_i|s_i, D)}{\pi_{\theta_0} (y_i|s_i, D)}]}
    + \mathbb{1}\times(i=n) \times r^M
\end{split}
\end{equation*}

where $\beta_{KL}$ is the KL coefficient, $n$ is the final time step corresponding to the total number of sentences in the document $D$. 
For the intermediate time step $i$ ($i < n$), the reward is the negative KL divergence between the output distribution of the backbone model and the current policy. For the final time step $i$ ($i = n$), when the model obtains the complete summary $S$, the reward model takes the summary and the original document to produce the final reward $r^M$ (see Section \ref{Reward}). 


\section{Interactive Learning with Online RL}
\begin{figure}[t]
    \centering
    \includegraphics[width=0.45\textwidth]{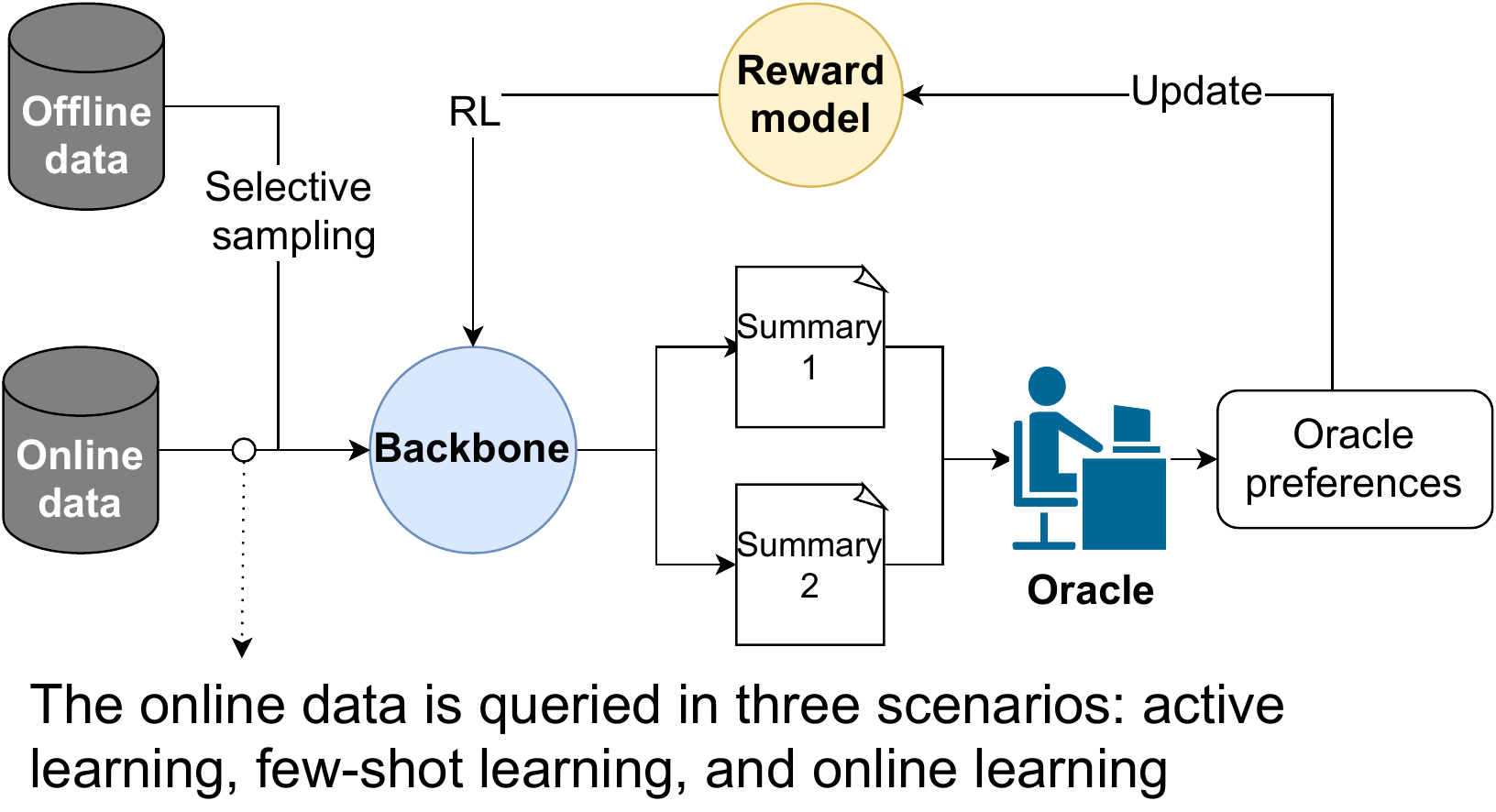}
    \caption{The overview of our framework. in which the backbone is in charge of generating two summaries for a document. Then the oracle selects which summary is better for a given document. The reward model afterward transforms the oracle's preference into a discrete signal to optimize the backbone. Our framework contains two novel components: efficient sampling from offline data and the preference-guided reward model.}
    \label{fig:overview}
\end{figure}
\subsection{Problem formulation}
In our problem, a backbone summarizer and a reward model are pretrained with \textit{offline} data by using supervised learning. We aim to update the models to adapt with a stream of \textit{online} data by using RL. Given a document $D$ from the online data, the backbone (Section \ref{backbone}) extracts two summaries for an oracle (i.e, a humans) to provide preference feedback. Each time the backbone receives feedback in one interaction. After receiving feedback from the oracle, the reward model is fine-tuned by using supervised learning (Section \ref{Reward}). The feedback is then translated to a reward via a reward model, which is later used to fine-tune the backbone by using RL (Section \ref{FRL}), thereby improving the summary quality for future interactions. 

This interactive mechanism is applicable to three scenarios: active, online, and few-shot learning (Section \ref{sec:interaction-plot}). Offline data can be employed during online fine-tuning to speed up the learning process by using the selective sampling method (Section \ref{sec:seclective_sampling}). The common objective of all scenarios is to maximize summarization quality while minimizing the number of interactions. We describe the overview of our framework in Figure \ref{fig:overview}.


\subsection{Interaction plots}\label{sec:interaction-plot}

\paragraph{Active learning}


Given a pool of unlabeled samples, the agent queries the most informative samples at each iteration to create summaries. Then the agent asks the oracle to select which summary is better for the queried documents. The agent learns from the feedback of the oracle subsequently. This setting has been studied by \citet{gao2018april}.

\paragraph{Online learning}
Different from active learning, for online learning, each unlabeled instance is typically drawn one at a time from the data stream, continuously being sent to the system. The agent processes the data and is not allowed to choose the document. \citet{ziegler2019fine} is the early work that promotes the online learning scenario. 

\paragraph{Few-shot learning}
Similar to the online setting, the system passively receives samples from the data stream. This time, it is only given a few unlabeled samples (up to 4 documents) and manages to learn from the data by continuously producing summaries and receiving feedback. To our best knowledge, this is the first time that the few-shot setting is examined in preference learning. 

\subsection{Efficient interaction from offline data}\label{sec:seclective_sampling}
To achieve sample efficiency in terms of the number of interactions between the backbone and the oracle, we argue that it is critical to leverage prior data used in the pretraining phase. We observe that the naive use of prior data by using random sampling is inefficient. Hence, we propose two novel offline-data sampling methods: low-reward and document-similarity. We describe all sampling mechanisms as follows.

\paragraph{Random sampling (Random)} We randomly select $k$ offline documents and combine them with online data to fine-tune the backbone model.

\paragraph{Low-reward sampling (LRS)} Intuitively, high-quality summaries should be given a higher reward by the reward model. Samples that have extracted summaries with low rewards have not been well-learned by the summarizer. As the result, we select top $k$ documents having the lowest rewards from the reward model.

\paragraph{Document-similarity sampling (DSS)} Conventionally, the backbone outputs good summaries when the training and testing distributions are similar. Therefore, we sample offline documents that share similar semantics with online documents. To do that, we encode the documents by using BERT \cite{devlin2018bert}, and then compute their similarity by using the Cosine distance. $k$ offline documents have the lowest distance to the online counterparts are selected for online training.


    



\section{The Reward Model for Preference Learning}\label{Reward}
\subsection{Training procedure}

We construct a reward model to generate preference rewards for finetuning the backbone summarizer with RL. The reward model should simulate human preferences, capable of assigning a higher reward to the preferred summary. For preference learning, we use the relative order between two summary candidates to train the reward model. The correct order is given by humans via their preferences. 

For traditional learning, the model has access to training documents of the dataset. Each document $D$ has a gold summary $S_g$, which is assumed to be preferred to other candidate summaries (silver or machine-generated $S_m$), denoted by $S_g\succ S_m$. The training data is in the form of a triplet ($D, S_g, S_m$) where $S_g\succ S_m$. We argue that this is insufficient to achieve a good reward model. From our observation, before reaching the quality of gold summaries, the machine often produces off-topic and too-short answers. In addition, the work of \citet{maxwell2017study} also suggested that the oracle prefers longer summaries because they feel that longer summaries are more readable, clear, and informative. Thus, we introduce three objectives as follows. 

\paragraph{The topic objective} Aiming to detect the document's topic: ($D, S_g, S_m$). We generate the training set of the \textbf{topic objective} as follows. For each document $D$ that has a gold summary $S_g$, we select a summary $S'_g$ from a different document $D'$ (randomly chosen) and expect that $S_g \succ S'_g$. If we miss the human summary, we could use machine-generated summaries instead (($D, S_m, S'_m$) such that $S_m \succ S'_m$).

\paragraph{The length objective} Aiming to detect the amount of summary information (i.e. the length of the summary). For each document $D$, we used a pretrained summarization model to generate a long ($S_{ml}$) and a short ($S_{ms}$) summary to obtain the triplet ($D, S_{ml}, S_{ms}$). The correct order is $S_{ml}\succ S_{ms}$.

\paragraph{The quality objective} Aiming to detect the general quality of a summary. For each document $D$, we generate a summary $S_m$ which has a similar length to the human summary and include the gold summary to compose the triplet for training ($D, S_g, S_m$). We note that the data of this objective is similar to the traditional data used in \citet{stiennon2020learning}, which needs humans to provide a gold-silver summary or correct order between arbitrary summary candidates. On the contrary, the data of \textbf{topic} and \textbf{length} objectives can be created programmatically and thus save human effort. 

\subsection{Respective-order mapping}

For summarization, we argue that the reward model's interpretation of ordering should be determined based on the distance between the summary and document representations. Therefore, we aim to learn global representing mapping $\phi$ such that:
\begin{equation*}
S_1\succ S_2 \iff d(\phi(D), \phi(S_1)) < d(\phi(D), \phi(S_2))
\end{equation*}
where $d()$ is the Euclidean distance, $\phi(D)$ is the representation of the document $D$ in the respective-order mapping. Then, the  reward for a summary $S$ is computed as follows.
$$r^M(D, S) = \frac{score(\phi(D), \phi(S)) - score_{min}}{score_{max}-score_{min}}$$


where $score(D, S)=\frac{1}{1+exp(d(\phi(D), \phi(S)))}$ is the unnormalized score between $D$ and $S$; $score_{min}$ and $score_{max}$ are the minimum and maximum score over all pairs of document-summary, respectively. We name our reward model as ROMSR (Respective-Order Mapping Score Reward).

\subsection{Representation learning}

To train $\phi$, first, we embed all documents that appear in triplet sets. To obtain important features of a document, we use the joint-embedding of a Transformer (capture a general meaning of the document) \cite{reimers-2019-sentence-bert,devlin2018bert} and keyword feature vectors (capture important keywords in the document) \cite{sharma2019self}. We denote this transformation as $f(D)$. 

In fact, this feature vector $f(D)$ can be used to compare the similarity between two documents. However, it is not robust enough to satisfy the relative comparison from the triplets. Therefore, we learn a new embedding ($encode(f(D))$) that tries to follow all the triplet conditions. In order to be a good representation of $f(D)$, the embedding is also reinforced by the reconstruction loss of the autoencoder. The embedded representation is our target mapping $\phi(D) = encode(f(D))$.
The loss for the learning of the document representation $\mathcal{L}$ is the combination of the autoencoder reconstruction loss $\mathcal{L}_{AE}$ and the respective-order loss $\mathcal{L}_{RO}$ (the triplet loss) as follows.

\begin{equation*}
    \mathcal{L} = \mathcal{L}_{AE} + \mathcal{L}_{RO}
\end{equation*}

\begin{equation*}
    \mathcal{L}_{AE} = \sum_{D \in \mathbb{D} \cup \mathbb{S}} \|f(D) - decode(\phi(D))\|
\end{equation*}\vspace{-0.7cm}

\begin{equation*}
\begin{split}
\mathcal{L}_{RO} = \sum_{(D,S,S') \in T} \max(0, d(\phi(D),\phi(S)) \\ - d(\phi(D), \phi(S')) + \alpha)
\end{split}
\end{equation*}

where $\mathbb{D}, \mathbb{S}$ is the set of original/summarized documents, $T$  is the union of all triplet sets, and $\alpha$ is the margin hyperparameter controlling the stretch in the representation space \cite{yu2018correcting}.


\section{Experimental Setup}

\subsection{Datasets}
We use three benchmark summarization datasets for our evaluation. \textbf{BillSum} comprises 22,218 US bill and human-written summary pairs collected from the US Government Publishing Office \cite{kornilova2019billsum}. The data is split into 18,949 training and 3,269 testing bills. \textbf{Reddit TIFU} is an English dataset collected from Reddit. It contains 12,000 posts divided into TIFU-long and TIFU-short (documents have less than 400 words) \cite{kodaira2018rule}. \textbf{Livedoor} contains Japanese articles crawled from the Livedoor News website. Each article consists of three summary sentences written by editors \cite{kim2018abstractive}.

Our experiments use the training set to pretrain the backbone and reward models. Then, we randomly select 5000 samples from the original training data to create offline data. The online data is created from the original validation data. For active learning and online learning settings, we sample up to 320 documents from the online data. For few-shot learning, only 4 documents are randomly selected as online data. We aim to minimize the number of interactions to reach the highest performance on the whole online dataset.

\subsection{Simulated interactions}
Given one original document with two corresponding generated summaries and a standard metric ROUGE score, the oracle knows the ground-truth summary, and prefers the summary that has the higher ROUGE score w.r.t the ground-truth. Theoretically, the ideal data for preference-based interactive learning is consistent preferences in which a higher-scored summary is always selected. However, we believe that perfect selection is impossible for real-world applications because humans can occasionally misinterpret the intention when the presented candidates have similar qualities. Therefore, in this work, we consider noisy preferences with uniform probability \(nc \in [0, 1]\), that randomly selects which summary is better.

\subsection{Baselines}
We construct a standard model (\textbf{baseline}) that just finetunes the backbone summarizer (see Section \ref{backbone}) with online data, coupled with our proposed reward model and PPO training. The combination of BERT and PPO is shown effectively in the offline document summarization task \citep{nguyen2021robust}. The baseline is similar to that of \citet{ziegler2019fine}, which uses human preferences and PPO to finetune language models. However, the latter uses a different reward model. 

To show the efficacy of our reward model, we compare ours with the reward model of \citet{ziegler2019fine} (\textbf{OpenAI}) and the uncertainty sampling reward model in \textbf{APRIL} \cite{gao2018april} in the preference prediction task (see Section \ref{sec:reward_model}).

 For few-shot and online learning, we compare our methods with the standard baseline. For active learning, we also compare our methods with APRIL. For all settings, we build a baseline that randomly samples from offline data to compare against our proposed sampling techniques: \textbf{LRS} and \textbf{DSS}. We note that all models share the same backbone and RL training. The baseline, Random, LRS, and DSS share the same reward model and only differ in the sampling techniques.


\subsection{Evaluation metrics}
We use ROUGE-scores \cite{lin2004rouge} for our evaluation, in which ROUGE-1 is the representative score. ROUGE-2 and ROUGE-L are reported in the Appendix. We also report the number of interactions to show that our novel sampling techniques can significantly speed up the agent's learning.

\section{Results and Discussion}

\subsection{Reward model study} \label{sec:reward_model}
We first observe the efficicy of the proposed reward model in the summarization process. To do that, we compare our reward model to the reward models of APRIL \cite{gao2018april} and OpenAI \citet{ziegler2019fine}. For the preference prediction task, the reward scores assigned to candidate summaries are used to determine the preferred summary. In general, the preferred summary is the one with a higher score. We select a subset of documents from the datasets to train the reward model. Each document can be used to construct three triples corresponding to the three objectives mentioned above. The number of pretraining/interactive training/testing documents is 1000/1000/1000 for Billsum and Livedoor, and 2000/2000/2000 for Reddit TIFU, respectively.




\begin{figure}[!h]
  \centering
  \includegraphics[width=0.48\textwidth]{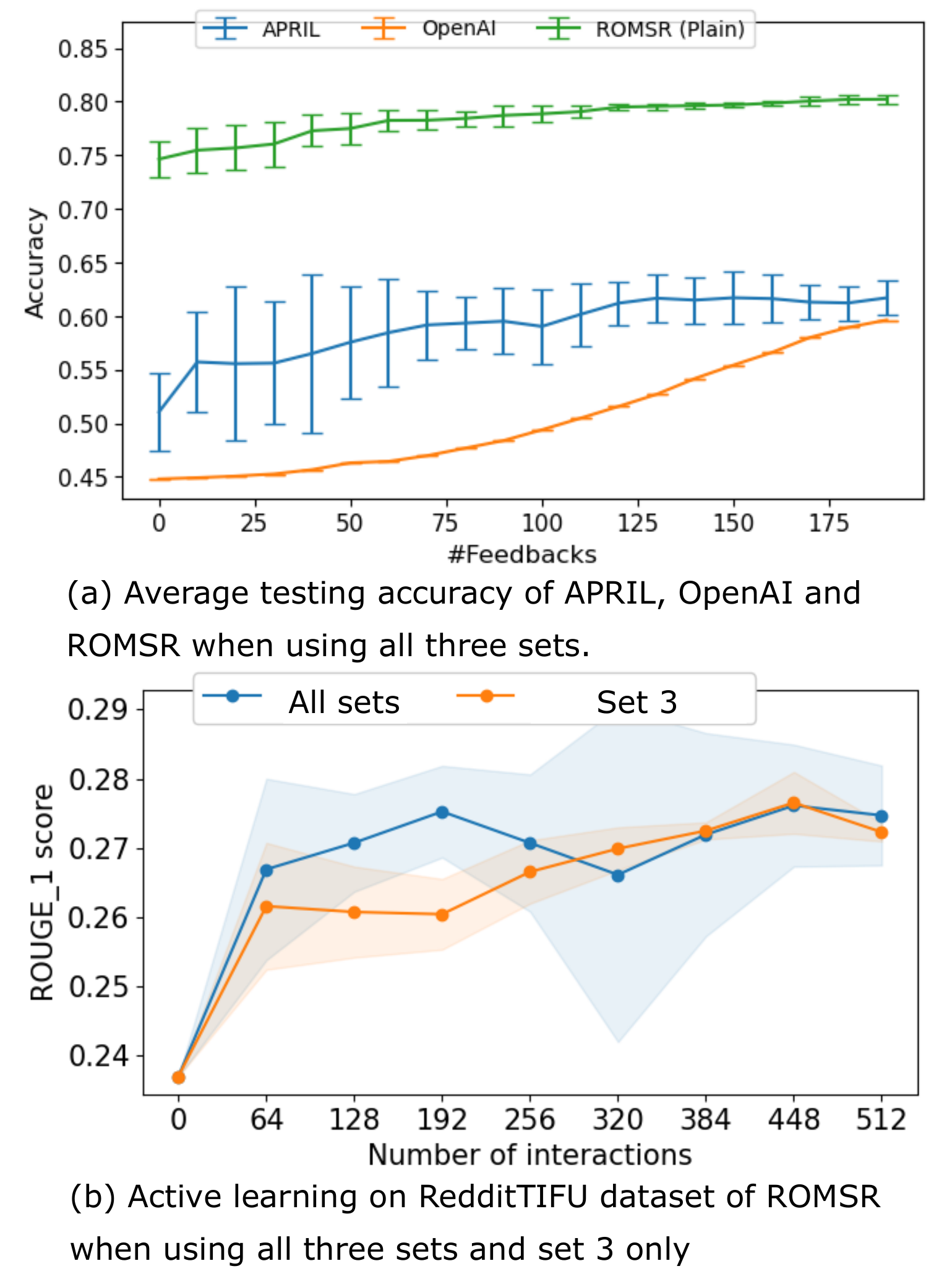}
  \caption{Reward model: (a) Accuracy of 3 models on all sets (b) The ROUGE-1 score of our ROMSR.} \label{fig:reward-compare}
\end{figure}

Figure \ref{fig:reward-compare} (a) demonstrates that our reward model (ROMSR) significantly outperforms other methods when training on all three objectives. APRIL uses a linear regression model combined with a heuristics function that is tuned for the quality objective. This specific design limits the model to handle all three objectives.
OpenAI's results exhibit slow learning progress because it is a large neural network using joint features from a pair of texts, while our reward model learns to use respective-order on a feature space of a single document. Moreover, the reconstruction loss from the autoencoder helps the model to avoid overfitting to small training samples. 

To verify the necessity of using three objectives for training the reward model, we compare two versions of our reward model: one is trained with all sets of objectives and another with the third objective (the traditional quality objective). Two reward models are used in active learning to finetune the backbone. Figure \ref{fig:reward-compare}(b) shows that training on all three objectives converges faster.






\begin{figure*}[!h]
    \centering
    \includegraphics[width=\textwidth]{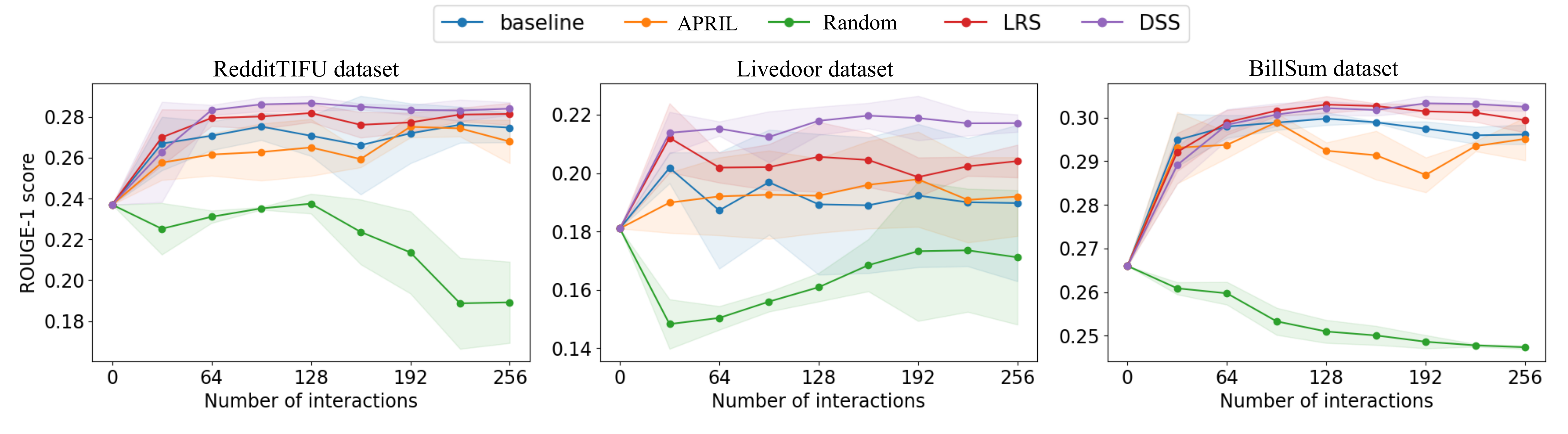}
    \caption{Active learning: ROUGE-1 with the mean and standard deviation over 5 runs.}
    \label{fig:active-res}
\end{figure*}

\subsection{Fine-tuning backbone model using human preferences} \label{sec:rl_result}
This section shows our comprehensive experiments on three datasets and three scenarios to prove the effectiveness of our human-preference-guided learning with RL, and our LRS and DSS sampling.

\paragraph{Active learning}



We compare our sampling techniques with the baselines, including random offline sampling and methods without offline sampling, such as the standard baseline and APRIL. Figure \ref{fig:active-res} shows that combining online data with randomly selected samples harms the agent during online fine-tuning. Meanwhile, selecting prior low-reward samples and similar documents boosts accuracy faster than not using offline data. After 64 interactions, ROUGE-1 reaches almost the highest. To validate the actual impact of ROMSR, we also test the reward model in the RL pipeline compared to the APRIL's reward. Due to the OpenAI reward is inferior to the APRIL's reward, we ignore the OpenAI's reward in this experiment. The results in Figure \ref{fig:active-res} show that APRIL (orange) is inferior to our ROMSR (blue) in all datasets.
\paragraph{Few-shot learning}

\begin{figure*}[!h]
  \centering
  \includegraphics[width=\textwidth]{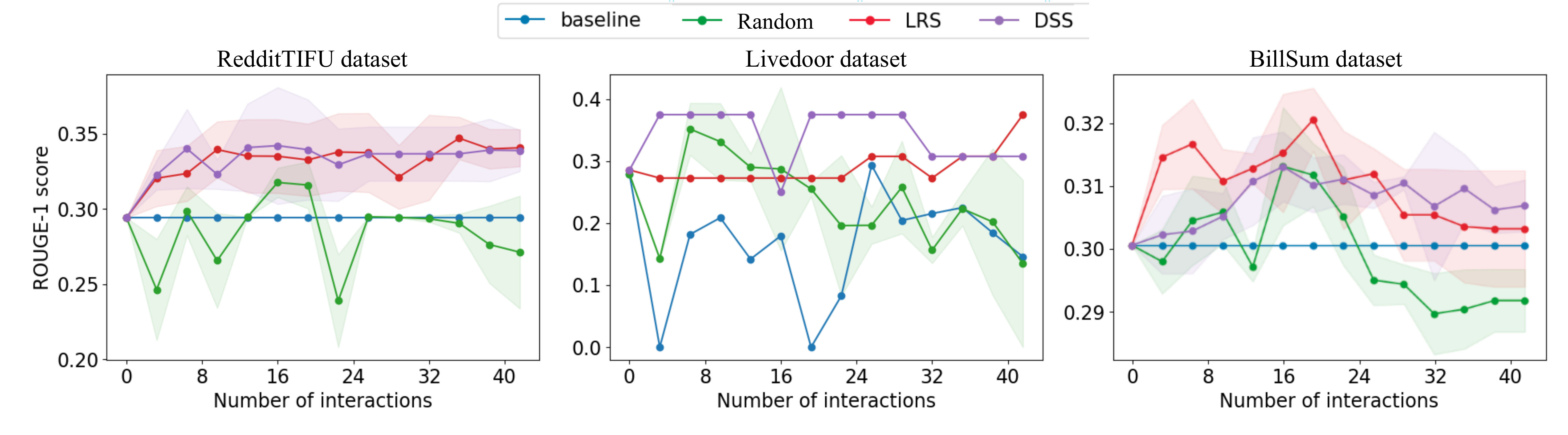}
  \caption{Fewshot learning: ROUGE-1 with the mean and standard deviation over 5 runs.}
  \label{fig:fs-res}
\end{figure*}

Figure \ref{fig:fs-res} describes the quality of the models: LRS, DSS, Random, and the standard baseline during interactions. The standard baseline without offline sampling can not improve the summarization agent with a few interactions and eventually worsen the agent. In contrast, LRS and DSS can significantly improve the model after 4 interactions. Naively using the offline data still harms the model, similar to active learning. 

\paragraph{Online learning}

\begin{figure}[!h]
  \centering
  \includegraphics[width=0.48\textwidth]{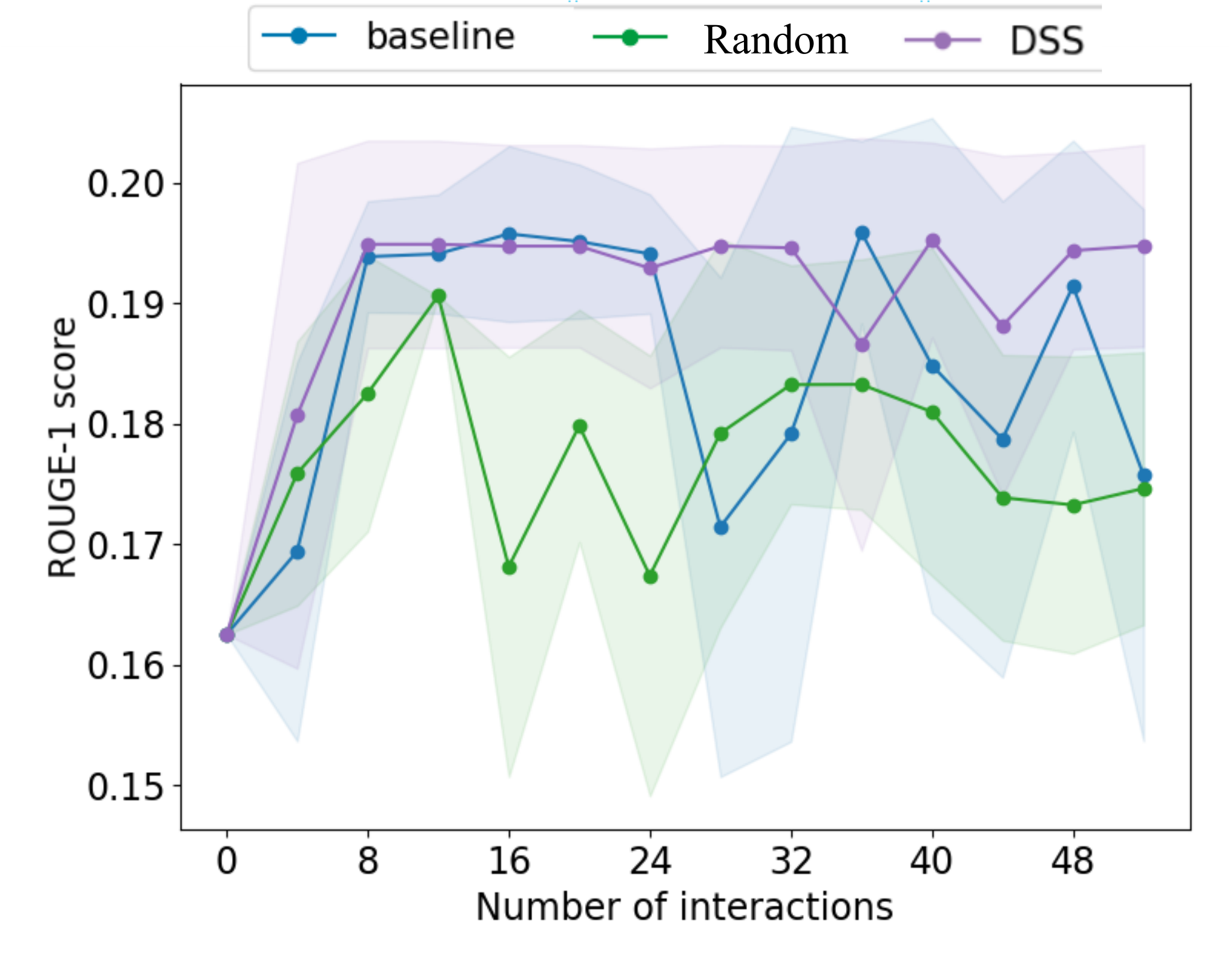}
  \caption{Online learning on Reddit TIFU: ROUGE-1 with the mean and standard deviation over 5 runs.}
  \label{fig:onl-tifu}\vspace{-0.4cm}
\end{figure}

 Due to computation limitations, we only examine this setting on the Reddit TIFU dataset. Also, prior experiments demonstrated that LRS and DSS are equally good sampling strategies. As DSS runs much faster than LRS, we choose DSS as our representative method in this experiment. We report the performance of DSS, Random, and the standard baseline in Figure \ref{fig:onl-tifu}.
 
The results show that random sampling is again inferior to other methods in this online scenario. The standard baseline shows good performance yet becomes unstable and drops performance later. Our DSS demonstrates fast and stable convergence, consistently achieving the highest ROUGE-1 score throughout interactions.

\begin{figure*}[htbp]
    \centering
    \includegraphics[width=\textwidth]{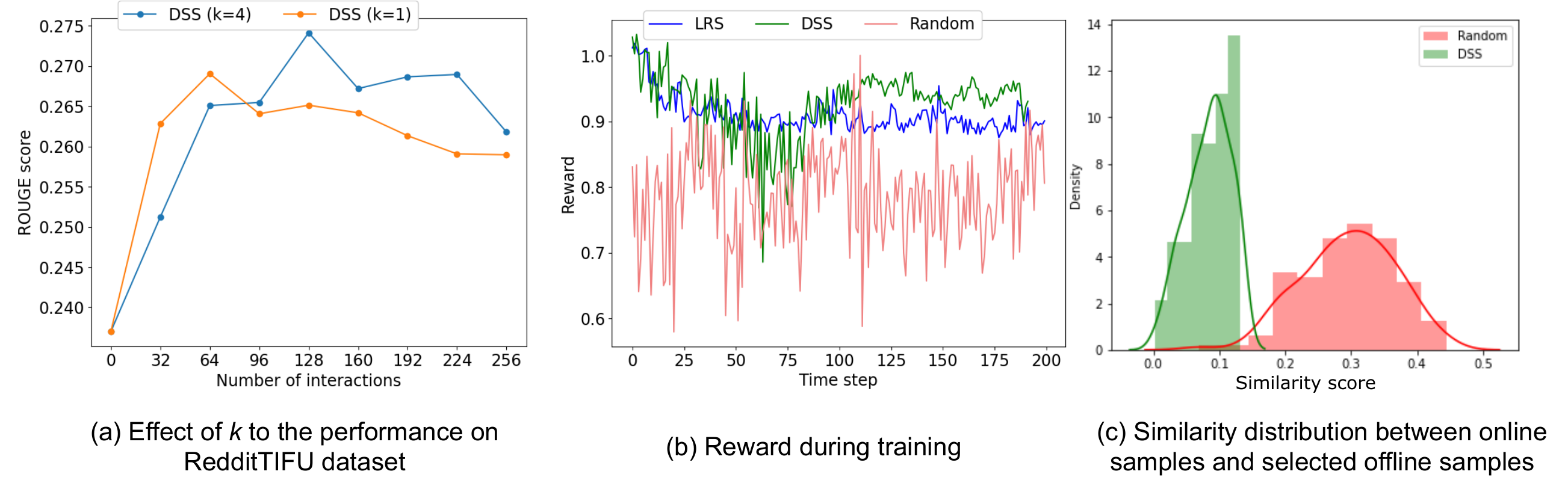}
    \caption{Ablation study: (a) Performance with different $k$ values; (b) Quality of selected samples; (c) Semantic similarity between online documents and offline documents.}
    \label{fig:abl_1}
\end{figure*}

\subsection{Ablation study and model analysis}




\paragraph{Impact of top-$k$ selected offline documents}
To investigate the effect of $k$-the number of sampled offline documents, we try different $k$ values ($k = 1$ and $k=4$) on the RedditTIFU dataset, and the results are depicted in Figure \ref{fig:abl_1}a. As observed, $k=4$ is slightly better, but $k=1$ runs and converges faster. Hence, we use  $k=1$ in all experiments. 

\paragraph{Random sampling makes low rewards}
We investigate the failure of random sampling. We keep track of the reward assigned to randomly selected offline documents across interactions. Figure \ref{fig:abl_1}b shows that, in general, DSS and LRS give better samples, indicated by higher assigned rewards, than random sampling. DSS and LRS's rewards are also much more consistent, showing less variance than random sampling. Thus, offline documents from random sampling provide little benefit and make the training unstable and unreliable. 

We also analyze the distribution characteristics of selected offline documents in Figure \ref{fig:abl_1}c. The green line represents the distribution of similarity scores between online and random offline samples. The red one is the distribution of similarity scores between online and DSS samples. It is noticeable that the similarity score of DSS is in the range from 0.0 to 0.15, whereas this score of Random is up to 0.5. It is expected that sampling offline data will create a coherent training distribution for online learning, but Random skews the training distribution. Therefore, Random harms the performance.

\begin{table}[htbp]
    \centering
    \begin{tabular}{c c c c} \hline
 Method & Interactions & Run time & ROUGE-1 \\ [0.5ex]
 \hline
 \multicolumn{4}{c}{The Reddit TIFU dataset} \\
 \hline
 Baseline & 64 & 2h20m & 26.0 \\ 
 DSS & 64 & 2h50m &  28.6 \\
 \hline
 \multicolumn{4}{c}{The BillSum dataset} \\
 \hline
 Baseline & 96 & 50m & 29.4 \\ 
 DSS & 96 & 1h40m &  30.3 \\\hline
\end{tabular}
\caption{The running time of active learning with offline sampling.}\label{table:running_time}
\end{table}

\paragraph{Running time}
We assess the running time of our proposed method compared to the baseline on Reddit TIFU and BillSum in Table \ref{table:running_time}. The assessment is conducted on a single Tesla T4. On Reddit TIFU, our DSS takes nearly 3 hours (64 interactions) to reach almost the highest ROUGE-1 score of 28.6, while the baseline reaches only 26.0 of ROUGE-1 after 64 interactions. The baseline takes only 30m faster than the DSS. In light of BillSum, the DSS is slower than the baseline, about 50 minutes for 96 interactions. However, the DSS is better than the baseline in terms of ROUGE-1. Despite that our proposed method takes more computing resources than the baseline, our approach requires much fewer interactions, saving human resources in the real HITL setting. In terms of real-time interactive feedback, it takes around 3 minutes for training in each interaction on the Reddit dataset, and approximately 1 minute for training in each interaction in the BillSum dataset, which is not really real-time. It is noted that the amount of time to provide feedback is not included because we use simulated oracles. Hence, in actual applications, we can fine-tune the model while the oracle gives feedback because it takes time for the oracle to read the summary and decide their preference. We believe such a parallel process will make the methods feasible to use in real-time settings. 



\section{Conclusion}
This work proposes a novel approach for learning reward functions in preference learning. By utilizing offline data with a reward model that focuses on the distance between the summary and the document, our method improves the ROUGE-scores by $2-5\%$ in comparison to the APRIL framework, while beating the random baseline by a large margin on three different datasets. Our experimental results also suggest that by applying low-reward sampling or document-similarity sampling, we can achieve efficiency in terms of both running time and the number of human interactions. Regarding limitations, our method is not tested with large online stream data, which may cause catastrophic forgetting. Future work will confirm our model's effectiveness for abstractive summarization.

\section*{Acknowledgement}
We would like to thank anonymous ACL ARR reviewers and Senior Area Chairs who gave constructive comments for our paper.

\bibliographystyle{acl_natbib}
\bibliography{references}

\clearpage
\appendix
\section{Hyper-parameters and Implementation details}
Due to the limitation of resources, our backbone bases on BERT-base with 12 Transformer blocks with the hidden size is 768, and 110M parameters. We train our model with the learning rate of 1e-5. During online fine-tuning with RL, we set $k$ equals the number of online documents, where $k$ is the number of selected offline data. The noisy peference probability $nc$ is 0.1.
For evaluation, we use ROUGE-score with parameters $-c\ 95\ -m\ -r\ 1000\ -n\ 2$.

\section{Additional Results}
\begin{figure}[!ht]
    \centering
    \includegraphics[width=0.8\textwidth]{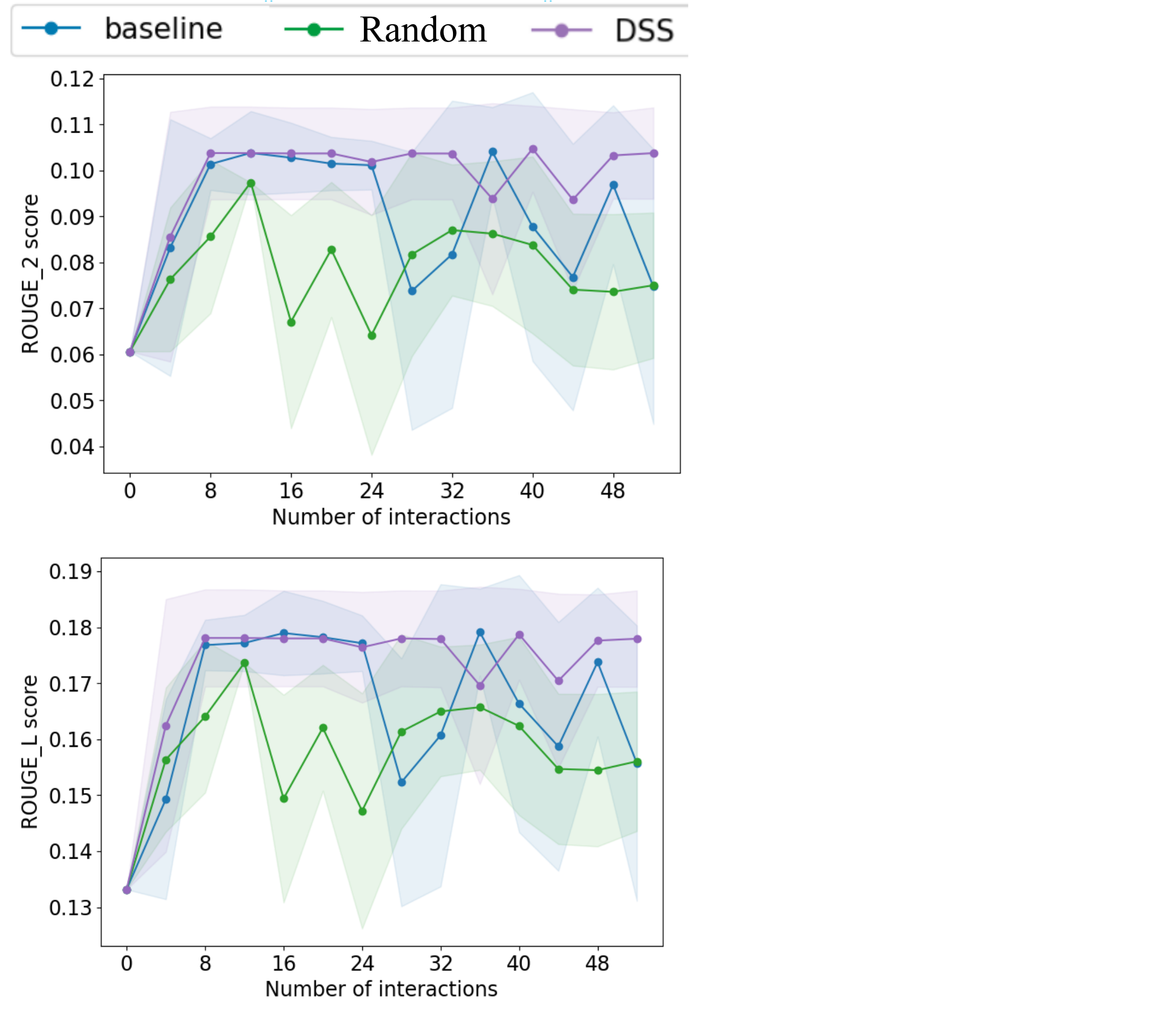}
    \caption{Online learning on RedditTIFU dataset: ROUGE-2 and ROUGE-L score with the mean and std over 5 runs}
    \label{fig:online-app}
\end{figure}

\begin{figure*}[t]
    \centering
    \includegraphics[width=\textwidth]{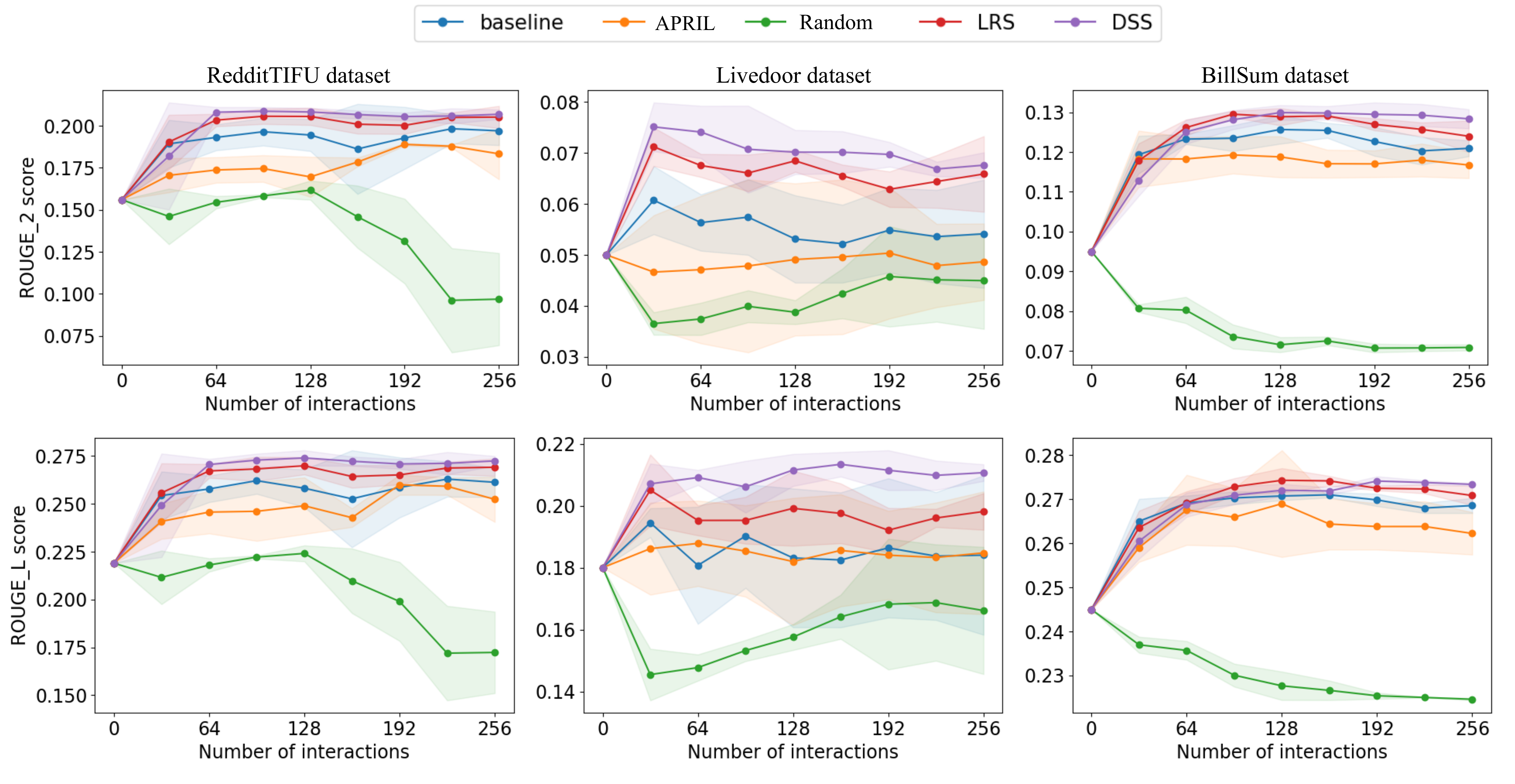}
    \caption{Active learning: ROUGE-2 and ROUGE-L score with the mean and std over 5 runs}
    \label{fig:active-app}
\end{figure*}

\begin{figure*}[t]
    \centering
    \includegraphics[width=\textwidth]{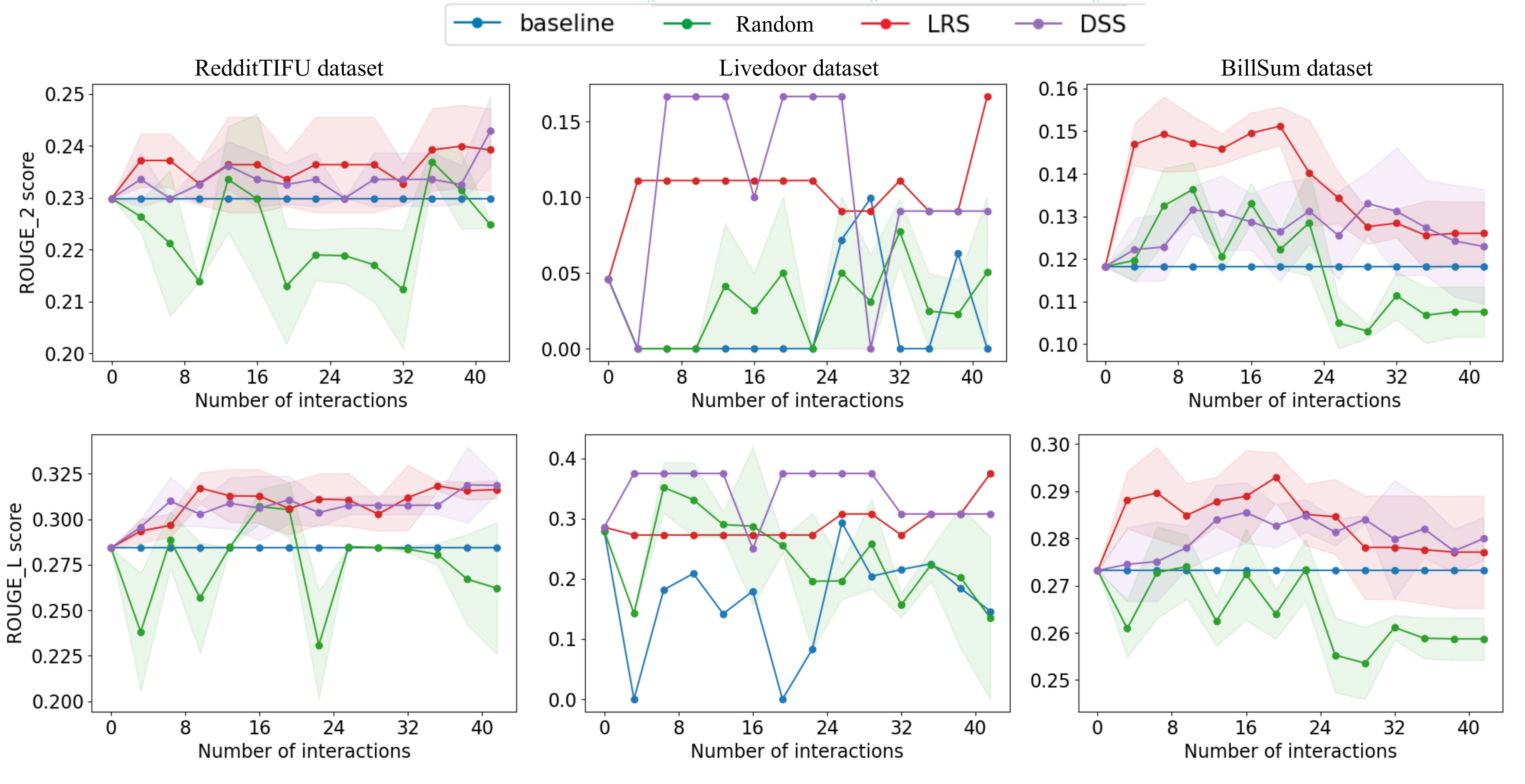}
    \caption{Fewshot learning: ROUGE-2 and ROUGE-L score with the mean and std over 5 runs}
    \label{fig:fewshot-app}
\end{figure*}

In Figure \ref{fig:active-app}, we show ROUGE-2 and ROUGE-L scores with the same setting as section \ref{sec:rl_result}. The results of few-shot and online learning setting is depicted in Figure \ref{fig:fewshot-app} and \ref{fig:online-app} respectively.

\end{document}